\documentclass[a4paper]{article}
\usepackage{spconf,amsmath,graphicx}
\usepackage{blindtext}
\usepackage{booktabs,tabularx}
\usepackage{subcaption}
\usepackage{caption}
\usepackage[english]{babel}

\title{Image Denoising and Super-Resolution using Residual Learning of Deep Convolutional Network}

\begin{document}

\maketitle
\begin{abstract}
Image super-resolution and denoising are two important tasks in image processing that can lead to improvement in image quality. Image super-resolution is the task of mapping a low resolution image to a high resolution image whereas denoising is the task of learning a clean image from a noisy input. We propose and train a single deep learning network that we term as SuRDCNN (super-resolution and denoising convolutional neural network), to perform these two tasks simultaneously . Our model nearly replicates the architecture of existing state-of-the-art deep learning models for super-resolution and denoising. We use the proven strategy of residual learning, as supported by state-of-the-art networks in this domain. Our trained SuRDCNN is capable of super-resolving image in the presence of Gaussian noise, Poisson noise or any random combination of both of these noises.   
\end{abstract}
\section{Introduction}
\label{sec:intro}

Image super-resolution and denoising are topics of great interest.
Image super-resolution is the task of obtaining a high-resolution image from a low-resolution image. It is an ill posed problem since multiple solutions can exist for a single low-resolution pixel. Initially interpolation-based image super-resolution methods were used since they were extremely simple and fast. Such methods perform dealiasing of the low-resolution image to up-sample. However this introduces severe  blurring effects due to which such methods gave sub-optimal results in restoring fine texture details. 

Sparse coding based methods are also widely used for super-resolution. The method involves several steps in its solution pipeline. However, not all steps in the pipeline are optimized. It has been shown in  \cite{dong2016image} the pipeline used in sparse coding based methods is equivalent to a deep convolutional neural network. This allows optimization of the entire pipeline.

The goal of image denoising is to recover a clean image {\it I} from a noisy observation {\it K} which are related as {\it K = I + V} where {\it V} is the noise (generally assumed to be additive white Gaussian noise).
The commonly used metric to measure noise in an image is the Peak-signal-to-noise ratio(PSNR) which is the ratio between the maximum possible power of a signal and the power of corrupting noise.
In the past, several models have been used for image denoising such as non-local self-similarity(NSS) models, sparse models, gradient models and Markov random field models. Though these methods give high performance in denoising, they are computationally expensive since they involve a complex optimization problem in the testing stage. 

Discriminative learning methods are being used recently to overcome this disadvantage. In \cite{zhang2017beyond}, Zhang et al. use discriminative learning methods to separate noise from a noisy image using a Convolutional Neural Network(CNN). Residual learning of CNN is being widely used now-a-days to solve the problem of performance degradation in deep networks. In residual learning the network learns a mapping from the input image to noise.   

In our work, we develop a single network capable of performing image super-resolution and denoising. We use the method of residual learning. Instead of learning end-to-end mapping we train the network to output a residual image. This residual image is difference of input image and the original image. The aim of this system is to remove noise and then enhance image resolution (i.e. super-resolution)
We name the proposed model super-resolution denoising convolutional neural network (SuRDCNN). The proposed CNN though deep enough for our task at hand, is quite simple in architecture. We have intentionally kept small convolution kernels and repetitive layer design. This structure gives it capability to learn different noise types of varying level. Moreover, in some aspects the performance of our network is better than some other deep learning models available for similar task. It is worth mentioning that when we compare our model with the models proposed earlier we use same training data and same training time. We trained our network on data set that not only contained patches from different images but also each patch had different (randomly chosen) noise level and noise type.


\section{Related Work}
\label{sec:relatedwork}

In \cite{dong2016image}, Dong et al. propose a deep convolutional neural network for super-resolution (SRCNN). Their network learns an end-to-end mapping from low resolution image to a high resolution image. They produce state-of-the-art results using a lightweight network consisting of 3 layers. Their method achieves fast speed and hence is suitable for online usage. 

\begin{enumerate}
    \item Patch extraction from a low resolution image and representation as a high dimensional vector.
    \item Non-linear mapping from a high dimensional vector to another high dimensional vector.
    \item Reconstruction of high resolution image.
\end{enumerate}

They used a kernels of 9 x 9, 1 x 1 and 5 x 5 in the 3 layers respectively. The number of feature maps in each of the three layers were 64, 32 and 3. 

In \cite{kim2016accurate}, Kim et al.  propose an improved CNN model VDSR (Very Deep Super Resolution) that leverages residual learning and gives significant improvement over the results obtained by SRCNN.\cite{dong2016image} Contrary to the SRCNN which is a shallow network with just 3 convolution layers, VDSR uses 20 layers.

In \cite{zhang2017beyond}, Zhang et al. propose a plain discriminative learning model to remove noise from an image. They construct a feed-forward denoising convolutional neural network (DnCNN) to learn the residue and use batch normalization to speed up the training process as well as boost the denoising performance. Rather than being limited to additive white Gaussian noise (AWGN) at a certain noise level, their model is capable of performing Gaussian denoising with unknown noise level (i.e.,blind Gaussian denoising). The use of residual learning allows to train a single CNN model to tackle several tasks such as Gaussian denoising, single image super-resolution and JPEG image deblocking. The network they used consists of 17 layers(in case of white Gaussian noise) or 20 layers(in the case of blind Gaussian noise).The first layer is Conv+ReLU with 64 filters of size 3x3xc, where c is the number of channels. The second to second last layers are Conv+BN+ReLU with 64 filters of size 3x3x64. The last layer is Conv with c filters of size 3x3x64. The activation function used is ReLU.

\section{The proposed SuRDCNN model}
\label{sec:modelproposal}

In this section, we present the proposed super-resolution and denoising CNN model(SuRDCNN).

\subsection{Network Architecture}
\label{ssec:networkarchitecture}
The SuRDCNN is a 20 convolutional layer deep neural network. It consists of 10 activation layers and 18 batch normalization layer. The activation function used by us is {\bf tanh}. The loss function used is mean squared error. The kernel size is 3x3 and is the same for all the 20 convolutional layers. The number of feature maps given to the first convolutional layer is 3 corresponding to the RGB channels of the image. The number of feature maps given to layers 2 to 20 is 64. The final output of the network has 3 channels. All the weights are initialized from normal random distribution. 

With such an architecture, the total parameters of our network are 672,835 out of which 670,531 are trainable parameters and 2,304 are non-trainable parameters. During the training phase the input to the network is 32x32 patch of a noisy bi-cubic interpolated image and the target output is 32x32 residual image. The network is capable of working for any size of input image with the weights it learns during training. Number of trainable parameters for convolution layer does not depend on the input size. So after training the network for 32 x 32 size patches we transfer the weights to a network that can take desired size input image.

\begin{figure}[ht]
	\centering
    \includegraphics[width=1.0\columnwidth]{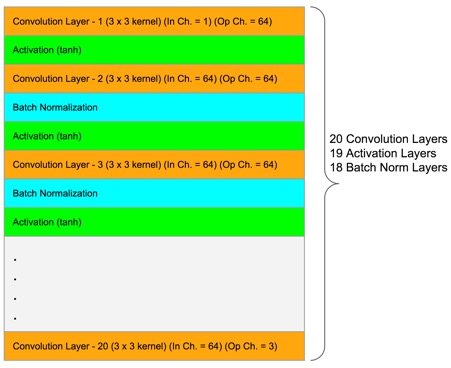}

\small
Figure 1: SuRDCNN Architecture\\
where In Ch. = Input Channels, Op Ch. = Output Channels
\end{figure}

\subsection{Training and Validation Data}
\label{ssec:data}
The dataset that we used to generate the training and validation data is the Berkeley Segmentation Dataset BSDS300. The dataset consists of 200 8-bit RGB images with pixel values from 0 to 255 for each of the three channels. The size of the images in the dataset is 481 x 321 or 321 x 481. In order to generate the training and validation data we scaled the pixel values from [0 255] to [0 1].

From each image we extracted 1000 random patches of size 32 x 32. Thus the total patches we generated are 200 x 1000 = 200,000 patches. Out of these 200,000 patches we used 160,00 for training and 40,000 for validation. Each such patch $P_{Original}$ is the desired output of the super-resolution and denoising system.

Now to create an input patch corresponding to each output patch, we scale the output patch to half the size in both dimensions(i.e. 16 x 16) to obtain $P_{Scaled}$. We then added noise to $P_{Scaled}$ to get $P_{Scaled+Noisy}$. We get the input to the network $P_{Input}$ by re-scaling $P_{Scaled+Noisy}$ to double size(i.e. 32 x 32) in both dimensions. We subtracted $P_{Original}$ from $P_{Input}$ to get $P_{Output}$(residual i.e. $P_{Output}$ = $P_{Input}$ - $P_{Original}$. Thus $P_{Input}$ is the input image corresponding to $P_{Output}$.

\subsection{Type of Noise}
\label{ssec:noisetype}
The following table contains information about the type and level of noise that we added to the input images.
\begin{center}
 \begin{tabular}{|| m{3em} | m{3cm} | m{3cm} ||} 
 \hline
 Type of Noise & Generation of Noise & Minimum PSNR \\
 \hline\hline
 Masked Gaussian Noise & Used MATLAB function imnoise() to add Gaussian white noise with mean 0 and randomly generated variance of the order 10-4. & approx. 34 dB (If maximum allowed level of masked Gaussian Noise is present) \\
 \hline
 Poisson Noise & Used MATLAB function imnoise() to add Poisson noise which uses input pixel values as means (scaled up by randomly generated factor of the order of 1010). & approx. 24 dB (If maximum allowed level of Poisson Noise is present)\\
 \hline
\end{tabular}
\captionof{table}{Type of Noise}
\end{center}

Minimum possible PSNR of input image is approx. 22 dB when both noises are at their maximum level

\section{Experiments and Results}
\label{sec:experimentsandresults}

\subsection{Experiments}
\label{ssec:experiments}

\subsubsection{Training our model}
\label{sssec:training}

The input to our network during the training phase are images(patches) of size 32 x 32. We used stochastic gradient descent as optimizer with a learning rate of 2 x $10^{-9}$. We trained the model for 50 epochs. The total training time was about 4 hours on 16 GB GPU Nvidia Quadro GP100.

\begin{figure}[ht]
	\centering
    \includegraphics[width=1.0\columnwidth]{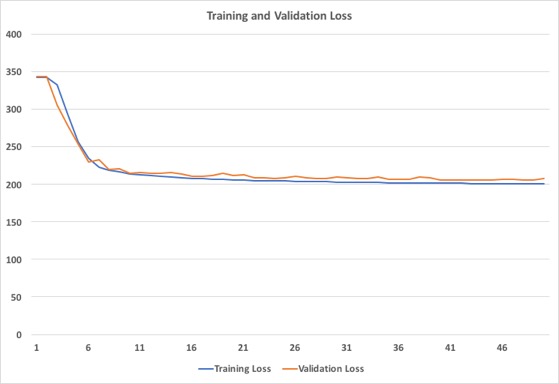}

\small
Figure 2: Plot of loss v/s Number of epochs
\end{figure}

\begin{figure}[ht]
	\centering
    \includegraphics[width=1.0\columnwidth]{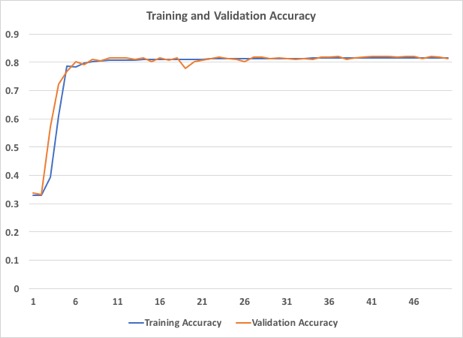}

\small
Figure 3: Plot of Accuracy v/s Number of epochs
\end{figure}

\subsubsection{Testing}
\label{sssec:testing}
In the testing phase suppose we are given a noisy image of size n x n, we interpolate it to an image of size 2n x 2n using bi-cubic interpolation. This noisy interpolated image is given to SuRDCNN which computes the residue of size 2n x 2n. We subtract this residue from the input image given to the network. This gives us our final super-resolved and denoised image.

\begin{figure}[ht]
	\centering
   \includegraphics[width=1.0\columnwidth]{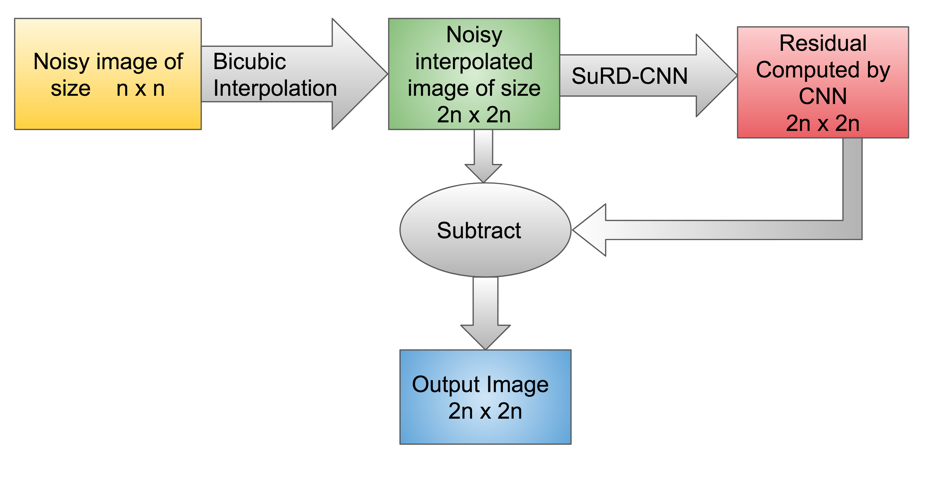}

\small
Figure 4: Generating output from SuRDCNN
\end{figure}

\subsection{Results}
\label{ssec:results}

\begin{center}

 \begin{tabular}{|| m{3em} | m{2cm} | m{2cm} ||} 
  \hline
 Images & Bi-cubic\newline(PSNR in dB) & SuRDCNN\newline(PSNR in dB) \\
 \hline\hline
 Plane & 27.91 & 28.25\\
 \hline
 Zebra & 20.96 & 22.06\\
 \hline
 Scenery & 21.33 & 23.13\\
 \hline
 Human & 24.21 & 24.53 \\
 \hline
 Tiger & 24.47 & 24.36 \\
 \hline
 Lizard & 23.04 & 24.19 \\
 \hline
\end{tabular}
\captionof{table}{Results}\label{table2}
\end{center}

\begin{figure}
\centering

\begin{subfigure}[b]{1.0\linewidth}
\caption{Low resolution noisy image (Zerba)}
\includegraphics[width=\linewidth]{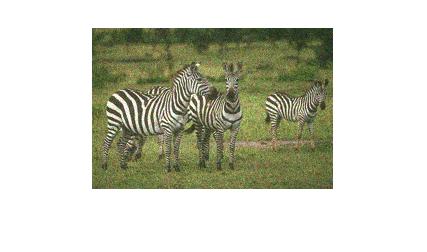}
\end{subfigure}

\begin{subfigure}[b]{1.0\linewidth}
\caption{Bi-cubic interpolation of image 'a' \newline
PSNR : 20.96 dB}
\includegraphics[width=\linewidth]{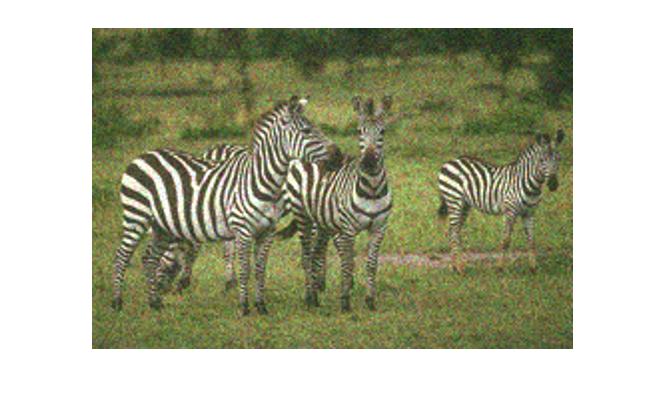}
\end{subfigure}

\begin{subfigure}[b]{1.0\linewidth}
\caption{Output of SuRDCNN system with 'b' as input \newline
PSNR : 22.06 dB}
\includegraphics[width=\linewidth]{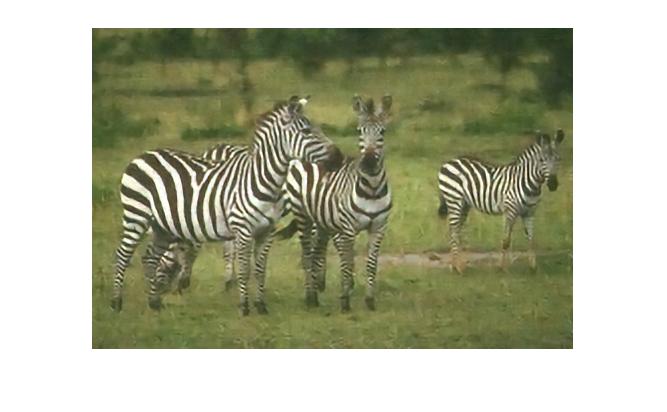}
\end{subfigure}

\begin{subfigure}[b]{1.0\linewidth}
\caption{Original high resolution image}
 \includegraphics[width=\linewidth]{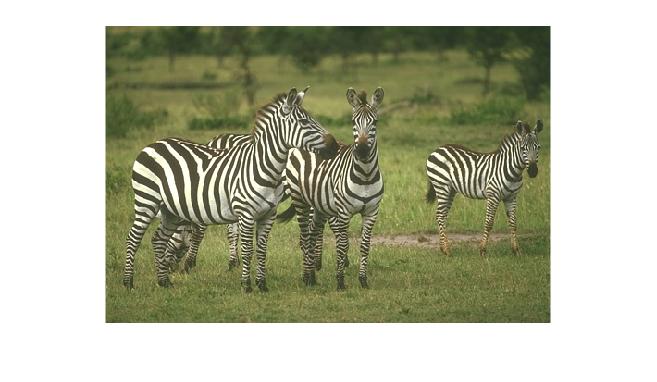}
\end{subfigure}
\end{figure}

\begin{figure}
\begin{subfigure}[b]{1.0\linewidth}
\caption{Low resolution noisy image (Scenery)}
\includegraphics[width=\linewidth]{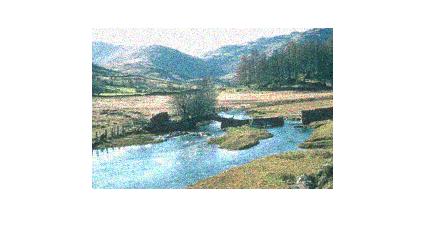}
\end{subfigure}

\begin{subfigure}[b]{1.0\linewidth}
\caption{Bi-cubic interpolation of image 'a' \newline
PSNR : 21.33 dB}
\includegraphics[width=\linewidth]{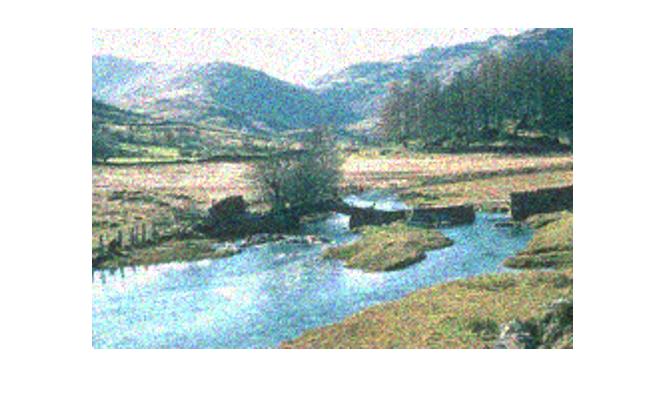}
\end{subfigure}

\begin{subfigure}[b]{1.0\linewidth}
\caption{Output of SuRDCNN system with 'b' as input \newline
PSNR : 23.13 dB}
\includegraphics[width=\linewidth]{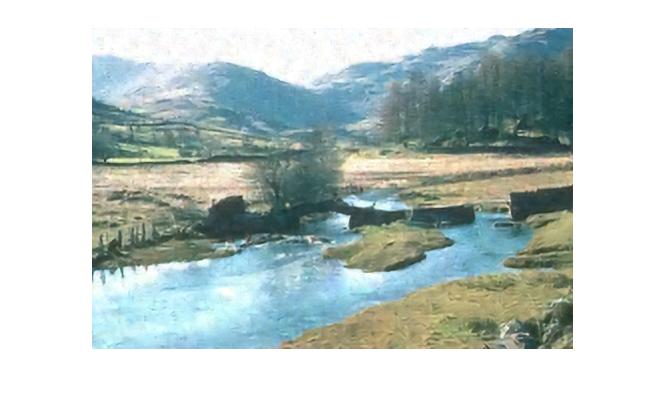}
\end{subfigure}

\begin{subfigure}[b]{1.0\linewidth}
\caption{Original high resolution image}
 \includegraphics[width=\linewidth]{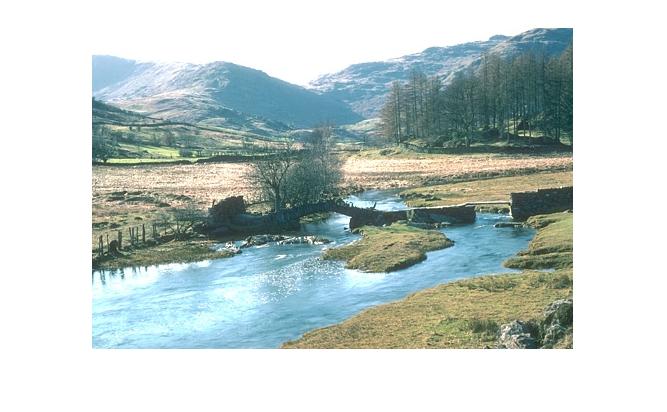}
\end{subfigure}
\end{figure}

We test the proposed SuRDCNN on different type of images with varying noise level. The test images contain Gaussian and Poisson noise of unknown level. Table 2 shows the PSNR (Peak-Signal-to-Noise-Ratio) values of few test images before and after processing the with SuRDCNN. We can see there is significant improvement in the PSNR values for all cases. Also improvement is quite evident even if noise level is not too high (say PSNR approx.28 dB). In such cases, we can assume that network is performing only super-resolution and no denoising. 
The quality improvement is so significant that we can easily perceive it visually.The images shown below the table substantiate this fact. 

Contrary to such good results by SuRDCNN, some existing popular models demonstrate complete failure at handling this task. One such popular model that fails completely is SRCNN. The super-resolution convolutional neural network (SRCNN) \cite{dong2016image} is presented with same noisy data (for which SuRDCNN) has been trained. But it is not able to learn any pattern for this data even after being trained for about 5 hours (approx. 600 epochs). 

\section{Conclusion}
We can conclude (from the results) that the network proposed is equally good for denoising, super-resolution and combination of both tasks. The network has sufficient depth to learn the pattern of noise from the data itself. A shallow network (like SRCNN) is not be able to perform this task because small number of layers (less number of trainable parameters) are insufficient to capture fluctuations due to noise level and type. Performing denoising task without prior information about noise requires learning noise from the image. Unless the network has a sufficiently large receptive field it will not be able to find pattern in the noise. In SuRDCNN this large receptive field is achieved by sufficiently large number of convolutional layers. Though each layer has a small convolution kernel but network overall has a large receptive field due to its depth.  Thus it seems likely that this network may capture other noise patterns apart from Gaussian and Poisson. Moreover, since this network has a promising run time and comparatively reasonable training time it stands a good chance of becoming a single point solution of image improvement in one shot.

\label{sec:references}


\end{document}